# Monte Carlo Matrix Inversion Policy Evaluation

Fletcher Lu*
School of Computer Science
University of Waterloo
Waterloo, Ontario, N2L 3G1, Canada

Dale Schuurmans
School of Computer Science
University of Waterloo
Waterloo, Ontario, N2L 3G1, Canada

## Abstract

In 1950, Forsythe and Leibler (1950) introduced a statistical technique for finding the inverse of a matrix by characterizing the elements of the matrix inverse as expected values of a sequence of random walks. Barto and Duff (1994) subsequently showed relations between this technique and standard dynamic programming and temporal differencing methods. The advantage of the Monte Carlo matrix inversion (MCMI) approach is that it scales better with respect to state-space size than alternative techniques. In this paper, we introduce an algorithm for performing reinforcement learning policy evaluation using MCMI. We demonstrate that MCMI possesses accuracy similar to a maximum likelihood model-based policy evaluation approach but avoids ML's slow execution time. In fact, we show that MCMI executes at a similar runtime to temporal differencing (TD). We then illustrate a least-squares generalization technique for scaling up MCMI to large state spaces. We compare this least-squares Monte Carlo matrix inversion (LS-MCMI) technique to the least-squares temporal differencing (LSTD) approach introduced by Bradtke and Barto (1996) demonstrating that both LS-MCMI and LSTD have similar runtime.

## 1 INTRODUCTION

Estimating the expected future reward in a Markov reward process is fundamental to many approaches for reinforcement learning and Markov decision process planning (Bellman, 1957). For instance, in Policy-Iteration (Howard, 1960), an estimate of the value of some current fixed policy must be performed at each iteration, a policy that is then improved with each iteration. In this paper we focus on value estimation. A variety of techniques for *performing value estimation* such as iterative successive approximations (eg. Jacobi iterative solvers) as well as reinforcement learning approaches such as temporal differencing (TD) and maximum likelihood (ML), exist. We will investigate the differing advantages between the latter reinforcement learning techniques and a Monte Carlo matrix inversion (MCMI) technique for solving a system of linear equations as the expected value of a statistic defined over a sampled random walk following a Markov process. In particular, we will investigate the problem of estimating the expected sum of future rewards in an infinite horizon discounted Markov reward process (Sutton & Barto, 1998).

* f2lu@cs.uwaterloo.ca

A popular approach to value estimation in Markov reward processes is by temporal differencing (Sutton, 1988; Dayan, 1992). Temporal differencing estimators are considered computationally efficient estimators that possess a bootstrap mechanism that allows for value estimate updates during the information gathering of sampling. Temporal difference estimation is known as a *direct* or model-free approach. This method does not require the explicit modelling of the state-to-state probability transitions or the average rewards. In contrast to such direct methods are indirect or model-based approaches such as the maximum likelihood method and the Monte Carlo matrix inversion method.

The maximum likelihood approach follows a model-based strategy for value estimation by forming an explicit estimate of the state transition matrix and average reward vector. The value estimate is derived by solving a matrix equation. The computational complexity of the matrix solve is often held as the main drawback of this approach which has been empirically shown to produce more accurate value estimates than model-free temporal differencing methods (Lu et al., 2002). However, Lu et al. (2002) have pointed out that ML's high complexity runtime is only a worst-case scenario arising rarely and often only with dense state transition matrices. In many cases, the state transition matrix is often quite sparse and there currently exist numerous direct matrix solvers to efficiently factor and produce solutions for such linear systems (Duff et al., 1986). However, there still remain sparse systems which result in the worst case cubic solve time for matrix solvers.

In contrast to the ML model-based method is another model-based method known as Monte Carlo matrix inversion (MCMI). Although MCMI was proposed over 50 years ago by J. von Neumann and S. M. Ulam in a paper by Forsythe and Leibler (1950), it is only recently that MCMI has been investigated in relation to reinforcement learning. Barto and Duff (1994) investigated the theoretical similarities between MCMI and iterative as well as dynamic programming based approaches to reinforcement learning. Their principal results dealt with how, due to the similarities to TD algorithms, TD algorithms should scale well for sufficiently large problems. They did not deal with implementing and using MCMI as a solver in its own right. In this paper we explicitly compare the implementations of MCMI with ML and TD. We also consider a least-squares generalization algorithm for MCMI to handle exponentially large states and compare it to the least-squares temporal differencing approach first introduced by Bradtke and Barto (1996).

MCMI differs from the TD method for finding a solution to a system of equations by casting the solution as an expected value of a statistic defined over a sequence of random walks. It is similar to the ML method in that both methods form an estimate of a matrix. However, ML estimates the original state transitions of the system, while MCMI estimates the inverse of a matrix system and then derives the value estimates directly from the reward vector by a matrix-vector product. An advantage of the MCMI model-based method over the ML method is that it avoids the costly matrix factorization that is needed to produce the solution in the maximum likelihood approach by deriving the inverse directly from sampling. Another significant advantage is that MCMI only requires storage space linear in the size of the state space versus ML's quadratic state space requirements.

It is important to distinguish MCMI with the Monte Carlo methods traditionally applied in reinforcement learning algorithms. The traditional Monte Carlo reinforcement algorithms are a form of iterative update similar to standard temporal differencing methods except they lack the bootstrapping which allows for immediate value estimate updates during sampling. They are therefore a form of model-free algorithm, unlike the model-based MCMI approach.

As noted by Barto and Duff (1994), MCMI can produce value estimates for a single state for a fixed policy without explicitly deriving the value estimates of other states. Unlike TD and ML whose value estimates are dependent on producing value estimates for all other states at the same time, MCMI essentially decouples this dependence on other states. This state value estimation independence will be a great advantage in our implementation of least-squares MCMI.

In this paper, we begin with a general overview of the fixed policy value estimation problem. This is followed by a description of the temporal differencing, maximum likelihood and Monte Carlo matrix inversion approaches to solving this problem, providing the theoretical background as to the advantages and limitations of each method. We then provide an algorithm for MCMI value estimation, detailing a runtime and estimation error analysis. We follow this with the introduction and analysis of a least-square Monte Carlo matrix inversion (LS-MCMI) method. We will then experimentally demonstrate that MCMI runs as efficiently as TD but produces significantly more accurate results just as other model-based methods such as ML do. We also demonstrate that MCMI tends to be only marginally less accurate than ML, but has the significant advantage that it scales much better in terms of state-space size than ML. Finally, we show that LS-MCMI and LSTD run at approximately similar rates.

## 2 BACKGROUND

The standard reinforcement learning environment, which we will be dealing with, involves a discrete time Markov reward process on a finite set of $N$ states, $n = 1, ..., N$ is described by a transition model $P(S_{i+1} = m | S_i = n)$, where we assume the transition probabilities do not change over process time $i$ (stationarity assumption). Such a transition model can be represented by an $N \times N$ matrix $P$, where $P(n, m)$ denotes $P(S_{i+1} = m | S_i = n)$ for all process times $i$. The reward $R_i$ observed at time $i$ is independent of all other rewards and states given the state $S_i$ visited at time $i$. We also assume the reward model is stationary and therefore let $r(n)$ denote $E[R_i | S_i = n]$ and $\sigma^2(n)$ denote $\text{Var}(R_i | S_i = n)$ for all process times $i$. Thus, $\mathbf{r}$ and $\boldsymbol{\sigma}^2$ represent the vectors (of size $N \times 1$) of expected rewards and reward variances respectively over the different states $n = 1, ..., N$.

The value function $\mathbf{v}(n)$ is defined to be the expected sum of discounted future rewards obtained by starting in a state $S_0 = n$. That is, $\mathbf{v}$ is a vector given by

$$\begin{aligned}\mathbf{v} &= \mathbf{r} + \gamma P \mathbf{r} + \gamma^2 P^2 \mathbf{r} + \cdots \\ &= \mathbf{r} + \gamma P \mathbf{v}\end{aligned} \quad (1)$$

Therefore, if $P$ and $\mathbf{r}$ are known then $\mathbf{v}$ can be calculated explicitly by solving the matrix equation

$$(I - \gamma P)\mathbf{v} = \mathbf{r} \quad (2)$$

All of the estimators we consider will produce estimates $\hat{\mathbf{v}}$ of the value function by processing sample trajectories that have been generated by some independent sampling strategy. The specific sampling strategy we consider depends on whether or not the Markov reward process has an absorbing state.

**Absorbing restarts** If the process has an absorbing state (and finite expected walk length) then the sampling process produces independent trajectories by restarting at a state randomly drawn from a uniform distribution whenever the absorbing state is reached.

**Random walk** If the Markov reward process does not have an absorbing state (and is irreducible) then we sample one long trajectory through the reward process.

Several estimators can be applied to the value estimation problem in Markov reward processes. These methods attempt to estimate the value of each state by processing sampled trajectories. The specific estimators we consider are: TD($\lambda$), maximum likelihood and Monte Carlo Matrix Inversion.

## 2.1 TEMPORAL DIFFERENCING

The temporal difference estimator TD($\lambda$) that we implement in our analysis is conventionally implemented using eligibility traces, as shown in Figure 1 (Sutton & Barto, 1998). TD($\lambda$) is perceived to be computationally efficient, as it runs in $O(TN)$ time, in the worst case, while requiring $O(N)$ space, where $N$ is the number of states and $T$ is the number of sampling steps.

## 2.2 MAXIMUM LIKELIHOOD

For the individual parameters $P(n,m)$ and $\mathbf{r}(n)$, the maximum likelihood estimates are given by

$$\hat{P}(n,m) = \frac{\#\{i : s_i = n \text{ and } s_{i+1} = m\}}{\#\{i : s_i = n\}}$$

$$\hat{\mathbf{r}}(n) = \frac{\sum_{\{i:s_i=n\}} r_i}{\#\{i : s_i = n\}}$$

if $\#\{i : s_i = n\} > 0$ (otherwise undefined). Here $\#$ denotes set cardinality. Given these quantities one can obtain the maximum likelihood estimate simply

```
Initialize $\hat{\mathbf{v}}_{TD}(n)$ arbitrarily, $\mathbf{e}(n) = 0$, $\forall\ 1 \leq n \leq N$
Repeat for each trajectory:
    Draw an initial state $n$
    Repeat for each step of trajectory:
        $a \leftarrow$ action given by $\pi$ for $n$
        Observe next state $m$ and reward $r$
        $\delta \leftarrow r + \gamma\,\hat{\mathbf{v}}_{TD}(m) - \hat{\mathbf{v}}_{TD}(n)$
        $\mathbf{e}(n) \leftarrow \mathbf{e}(n) + 1$
        For all states $\ell$:
            $\hat{\mathbf{v}}_{TD}(\ell) \leftarrow \hat{\mathbf{v}}_{TD}(\ell) + \alpha\,\delta\,\mathbf{e}(\ell)$
            $\mathbf{e}(\ell) \leftarrow \gamma\lambda\mathbf{e}(\ell)$
        $n \leftarrow m$
    Until state $n$ is terminal
```

Figure 1: On-line TD($\lambda$) with eligibility traces

by plugging $\hat{P}$ and $\hat{\mathbf{r}}$ into Equation 2 and solving for the vector $\hat{\mathbf{v}}_{ml}$ in

$$(I - \gamma \hat{P})\,\hat{\mathbf{v}}_{ml} = \hat{\mathbf{r}} \quad (3)$$

Solving this equation is perceived to be the most arduous aspect of producing an ML estimate, since it can require $O(N^3)$ run time using standard algorithms. Nevertheless, ML yields a consistent estimator in the sense that $\lim_{T \to \infty} \hat{\mathbf{v}}_{ml} \to \mathbf{v}$ with probability one for reachable states, since both $\hat{P} \to P$ and $\hat{\mathbf{r}} \to \mathbf{r}$ by the strong law of large numbers (Ash, 1972). However, ML is actually *biased*; that is, generally, $\mathrm{E}[\hat{\mathbf{v}}_{ml}] \neq \mathbf{v}$.[1] However, despite this bias ML yields a good estimator for $\mathbf{v}$ because it tends to make efficient use of the sample data by estimating the transition probabilities $\mathrm{P}(S_{i+1} = m | S_i = n)$ in terms of every visit to $S_i = n$ regardless of process time $i$. We empirically verify below that it does indeed yield superior estimates. In addition, ML requires $O(t)$ space where $t$ is the number of nonzeros in the matrix $I - \gamma \hat{P}$ of Equation 3.

## 2.3 MONTE CARLO MATRIX INVERSION

J. von Neumman and S. M. Ulam observed that for any matrix M where M's eigenvalues are bound by

---

[1] This can be seen by noting that even though $\mathrm{E}[\hat{P}^j \hat{\mathbf{r}}] = \mathrm{E}[\hat{P}^j]\mathrm{E}[\hat{\mathbf{r}}]$, $\mathrm{E}[\hat{\mathbf{r}}] = \mathbf{r}$ and $\mathrm{E}[\hat{P}] = P$ (since $R_i$ is independent of $S_k$ given $S_i = n$), it is not true that $\mathrm{E}[\hat{P}^j]$ is equal to $P^j$ in general. Consider the special case of determining $\mathrm{E}[\hat{P}^2(n,m)]$. Here $\hat{P}^2(n,m) = \sum_{\ell=0}^{N-1} \hat{P}(n,\ell)\hat{P}(\ell,m)$, where for terms such that $\ell \neq n$ we have $\hat{P}(n,\ell)$ independent of $\hat{P}(\ell,m)$, as desired. However for the term $\ell = n$, the quantities $\hat{P}(n,n)$ and $\hat{P}(n,m)$ are *not* independent. For example, in the case where $m = n$ they become $\hat{P}(n,n)^2$, whose expectation is given by $\mathrm{E}[\hat{P}(n,n)^2] = \mathrm{Var}(\hat{P}(n,n)) + (\mathrm{E}[\hat{P}(n,n)])^2 > (\mathrm{E}[\hat{P}(n,n)])^2$.

one $(\max_r |\lambda_r(M)| < 1)$, then

$$([I - M]^{-1})_{ij} = \sum_{k=0}^{\infty}(M^k)_{ij}. \quad (4)$$

From this observation, they were able to propose a mechanism for computing the individual elements $([I-M]^{-1})_{ij}$ of the inverse matrix $M$ by first splitting the matrix $M = P.*V$ where $.*$ represents the element by element dot product (ie. $M_{ij} = P_{ij} * V_{ij} \forall i,j$). $P$ is a probability transition matrix where $\sum_j P_{ij} = p_i < 1$ for each row of $P$. Wasow (1952) proposed a random walk where the walk starts in a state $i$ and the next state is determined by the probability distribution of row $i$ of matrix $P$. We move to next state $k$ from state $i$ with probability $P_{ik}$. The random walk continues in this way until a stopping point is reached. Recall that each row has $\sum_j P_{ij} = p_i < 1$. Therefore, for each row $i$, there is a probability of stopping of $1 - \sum_j P_{ij}$. Let $W_{ij}$ be the product

$$W_{ij} = \begin{cases} \frac{V_{i i_0} V_{i_0 i_1} \ldots V_{i_m k}}{1 - p_k} & \text{for } k = j \\ 0 & \text{otherwise} \end{cases} \quad (5)$$

where the sequence $\{i, i_0, i_1, ..., i_m, k\}$ are the resulting states of a single random walk starting in state $i$ and terminating in state $k$. Forsythe and Leibler showed that

$$E[W_{ij}] = ([I - M]^{-1})_{ij}. \quad (6)$$

An obvious solution to the value estimation of equation 2 is to find the inverse of $I - \gamma P$. For $\gamma < 1$, then the eigenvalues of $\gamma P$ are less than one. Therefore we can apply the Monte Carlo matrix inversion algorithm to find a value estimate to equation 2.

## 3 MONTE CARLO MATRIX INVERSION ALGORITHM

Finding the inverse of $I - \gamma P$ of equation 2 is well suited to Monte Carlo Matrix inversion because it satisfies the eigenvalue requirements and matrix $P$ is already a probability matrix. Note that the $\sum_j P_{ij} = 1$. However, as long as $\gamma < 1$ then $\sum_j (\gamma P)_{ij} = \gamma < 1$. Therefore the stopping value of our random walk trajectories will be $1 - \gamma$ at every step. Recall from equation 4 that if we are inverting $I - M$, we need to split $M$ into two matrices, $P'$ and $V$ where $\sum_j P'_{ij} < 1$ and $M_{ij} = P'_{ij} V_{ij}$. However, in our value estimation problem we can directly set $P' = \gamma P$. So, $V$ simplifies to a matrix of all ones. Therefore, in a random walk $W_{ij}$, equation 5 becomes

$$W_{ij} = \begin{cases} 1 & \text{for } k = j \\ 0 & \text{otherwise} \end{cases}, \quad (7)$$

```
Initialize column vectors t = 0, s = 0, v = 0,
set γ, fix policy π
and U is a uniform probability distribution {0,1}
Repeat for each trajectory:
    Draw an initial state n
    Repeat for each step of trajectory:
        t(n) ← t(n) + 1
        Choose x ∈ U
        While x ≤ γ and n is not an absorbing state repeat:
            Draw next state n
            t(n) ← t(n) + 1
            Choose x ∈ U
        s ← s + t
        v ← v + r(n)t
For each state n:
    v(n) ← v(n) / ((1-γ)s(n))
where r are rewards observed during trajectory sampling
```

Figure 2: Monte Carlo Matrix Inversion Value Estimation

where $i$ is the initial starting state and $k$ is the final state of our random walk trajectory. We can thus calculate any individual $([I-\gamma P]^{-1})_{ij}$ element by starting our trajectories always in state $i$ and the ratio of the number of states ending in state $j$ to the total number of trajectories would be the Monte Carlo matrix inversion estimate of $([I - \gamma P]^{-1})_{ij}$. We can calculate an entire row $i$ of matrix $[I - \gamma P]^{-1}$ by simply starting all our trajectory walks in state $i$ and storing the number of walks that end in each of the states $1 \leq j \leq N$. With an entire row $i$ of $(I - \gamma P)^{-1}$, we can calculate the value estimate for a single state $i$ by vector product

$$v_i = (I - \gamma P)^{-1}_{i*} \mathbf{r}. \quad (8)$$

One efficiency improvement that may be incorporated as noted by Forsythe and Leibler (1950) is to use each step of a trajectory as the start of a new trajectory, thus increasing the number of estimates derived from a single random walk to equal the number of steps in that walk. Figure 2 illustrates a Monte Carlo matrix inversion algorithm which produces values estimates for all visited states. Also, if an absorbing state is reached, the algorithm can simply stay in the absorbing state until a stopping $x$ value is obtained. However, it can be shown that the expected value of absorbing states is always $1/(1 - \gamma)$, so for added efficiency, random walks may be terminated and a new trajectory started when an absorbing state is reached.

### 3.1 ACCURACY ANALYSIS

As illustrated in equation 6, our random walk $W_{ij}$ has an expected value of $([I - \gamma P]^{-1})_{ij}$. However the ac-curacy of our solution will depend on the variance of our trajectory walks. The variance for our $[I - \gamma P]$ matrix can be shown to be

$$\sigma^2_{W_{ij}} = \frac{([I - \gamma P]^{-1})_{ij}}{1 - \gamma} - (([I - \gamma P]^{-1})_{ij})^2. \quad (9)$$

Since variance by definition is always positive, then the variance can be shown to be always bound by:

$$\sigma^2_{W_{ij}} \leq \frac{1}{4(1 - \gamma)^2}. \quad (10)$$

Therefore, error in Monte Carlo matrix inversion is dependent on both the value of $\gamma$ and the variance of the probability distributions of $P$. The smaller the value of $\gamma$, the better our estimates. This is in contrast to maximum likelihood, whose error in estimating $P$ is dependent only on the variance of the probability distribution of $P$ (Varga, 1962). Therefore we would expect that maximum likelihood value estimates should have less error than the Monte Carlo method. However, it would be superior to temporal differencing since it makes efficient use of sampling similar to maximum likelihood.

### 3.2 RUNTIME ANALYSIS

MCMI's runtime for a single estimate is determined completely by the stopping parameter $\gamma$. Since an estimate of $([I - \gamma P]^{-1})_{ij}$ can only be made at the end of each random walk, then value estimates can only be updated at the end of each random walk. The expected value of a random walk $W_{ij}$ is $\frac{\gamma}{1-\gamma}$ with variance $\frac{\gamma}{(1-\gamma)^2}$. Over a large number of random walks, the time until an update will be the expected value of a random walk, which can be treated as a constant. Therefore,

$$\text{number of walks} \propto T,$$

where $T$ is the total number of sampling steps. Following the algorithm of figure 2, where all states that are visited during a walk are updated at the end of a walk, then the total runtime of MCMI would be

$$\text{runtime} = \#\{\text{walks}\} \times \#\{\text{states updated}\},$$

which under worst case would be $\bullet(TN)$. This is the same runtime as temporal differencing. Therefore, we would expect MCMI to run at approximately the same rate as TD.

### 3.3 STORAGE COSTS

MCMI, similar to maximum likelihood, is a model-based method and therefore requires some model storage. However, unlike ML, MCMI has the advantage of only needing to store space linear in $N$, the number of states one is interested in value estimating.

---

Initialize column vectors $t = 0$, $s = 0$, $v_M = 0$,
$w = 0$, set $\Phi^{-1}$ and $\gamma$, fix policy $\pi$
and $U$ is a uniform probability distribution $\{0,1\}$
Repeat for each trajectory:
  Draw an initial state $n$
  Repeat for each step of trajectory:
    $t(n) \leftarrow t(n) + 1$
    Choose $x \in U$
    While $x \leq \gamma$ and $n$ is not an absorbing state repeat:
      Draw next state $n$
      $t(n) \leftarrow t(n) + 1$
      Choose $x \in U$
    $s \leftarrow s + t$
    $v_M \leftarrow v_M + r(n)t$
For each state $n$ of the $m$ states of set $M$:
  $v_M(n) \leftarrow \frac{v_M(n)}{(1-\gamma)s(n)}$
For each row $n$ of the k rows of $\Phi$:
  $w(n) \leftarrow \sum_{j \in M} \Phi^{-1}(n,j) v_M(j)$
where $r$ are rewards observed during trajectory sampling
and $M$ is the set of all states visited.

Figure 3: Least-Squares Monte Carlo Matrix Inversion (LS-MCMI) Value Estimation

## 4 GENERALIZATION

Thus far we have principlely dealt with tractable MDP state sizes. However in real world applications, state sizes are typically exponentially large. MCMI value estimation can deal with these large states in a manner similar to other reinforcement learning solvers through functional approximation. Functional approximators generally must be either known or available a priori to value estimation. The states are derived through some parameterized function where the parameter is a tractable size.

As an example of how we may apply a reinforcement learning generalization technique to MCMI let us consider one recent form of approximation using a least-squares technique. The basic idea presented by Bradtke and Barto (1996) is that an exponentially large state space can be compressed down into a smaller set under the assumption that some states behave similarly to other states. The compression is dependent on a set of feature vectors $\{\phi_1, ..., \phi_k\}$, where the number of feature vectors is a tractable size. This least-squares method modifies equation 2 by using a set of $k$ *feature vectors* $\phi_i$ each of size $N \times 1$ where $k \leq N$. These feature vectors capture similar properties among states. If we let $\Phi w = v$ where $\Phi = [\phi_1, ..., \phi_k]$, then equation 2 becomes

$$(I - \gamma P)\Phi w = r. \quad (11)$$

The matrix $\gamma P \Phi$ is an over-determined matrix. The Least-Squares Temporal Differencing approach

(LSTD) finds a solution by using a least-squares approach which multiplies both sides of equation 11 by the transpose of $\Phi$

$$\Phi^T(I - \gamma P)\Phi \mathbf{w} = \Phi^T \mathbf{r}. \qquad (12)$$

When performing LSTD, value estimates are computed for the $k$ feature vectors, which can then be mapped back to the original states.

For a least-squares Monte Carlo matrix inversion (LS-MCMI) approach, consider rearranging equation 11 such that:

$$\mathbf{w} = ((I - \gamma P)\Phi)^{-1}\mathbf{r}, \qquad (13)$$

which becomes

$$\mathbf{w} = \Phi^{-1}(I - \gamma P)^{-1}\mathbf{r}. \qquad (14)$$

Assuming that we can compute the inverse of the feature matrix $\Phi$, then we can modify our algorithm of figure 2 into the algorithm of figure 7.[2] Although the total number of states, $N$, in our system is exponentially large, the number of actual states visited, $m$, can be assumed to be a tractable size. Thus, if $M$ is the set of states visited, then $m = |M|$ and $m \ll N$. With only $m$ states visited, we can tractably store these in a value estimate vector $\mathbf{v}_M$ of size $m$ entries. We can then left multiply $\mathbf{v}_M$ with our $\Phi^{-1}$ to produce value estimates $\mathbf{w}$ of our $k$ feature states.

It is important to note that we can separate the process of estimating the value of actual states from the $k$ feature states of $\mathbf{w}$ only because MCMI has this uncoupled state independence property for the valuing of each state. Temporal differencing's state updates are dependent on the current value of all other states in the system. So for exponentially large MDPs, we can only get a good value estimate if we can update all states, which can only be tractably done if we deal with the compressed system $\Phi^T(I - \gamma P)\Phi$ at each update. Thus, MCMI has the significant advantage that we can produce value estimates for actual states of the original system and we only need to convert to the $k$ feature parameter states after completing our sampling.

In regards to runtime, we can replace $N$ in our analysis of MCMI with $m$ for LS-MCMI to yield a runtime of $O(Tm)$. The storage space requirement for LS-MCMI would be $O(m)$.

---

[2] Recall that in order to apply MCMI we need $\max_r |\lambda_r(I - \gamma P)| < 1$, which is true as long as $\gamma < 1$. This restriction is all we need for LS-MCMI as long as we only use MCMI for computing $(I - \gamma P)^{-1}$ and apply the $\Phi^{-1}$ transformation separately.

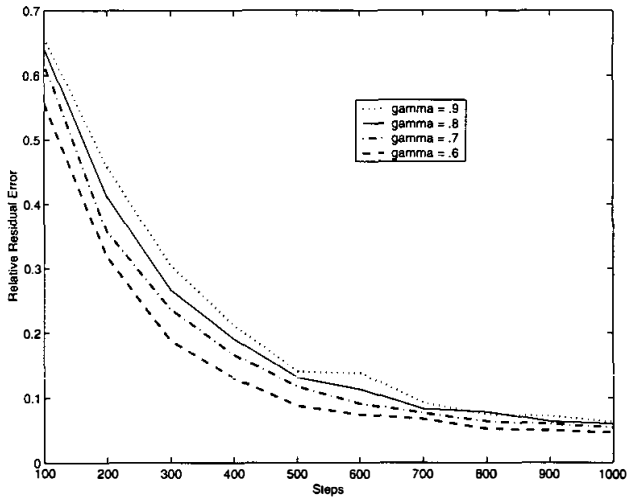

Figure 4: MCMI Error Reduction vs. Number of Sampling Steps (T) for differing $\gamma$, N = 300

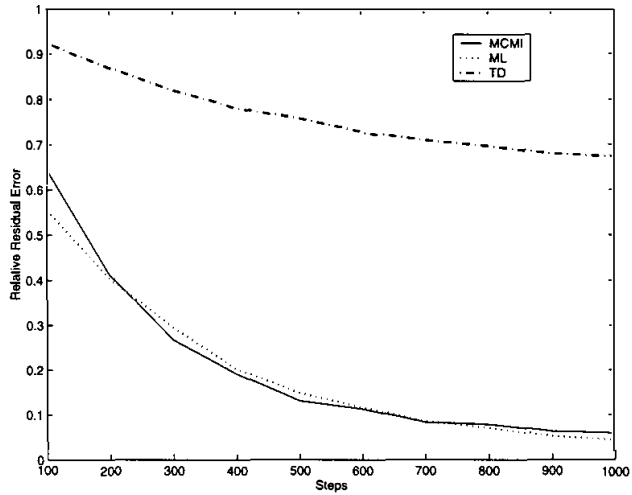

Figure 5: Error Reduction vs. Number of Sampling Steps (T) for differing solvers, N = 300

## 5 EXPERIMENTAL RESULTS

In our experiments, we artificially created random probability transition matrices. We therefore can compute a true value estimate for each state ($v_{true}$) for residual error comparisons. Unless otherwise noted, reported experimental results used $\lambda = .9$, $\alpha = .5$, $\gamma = .8$. The sample runs were each repeated 20 times and then the average reported.

Figure 4 demonstrates that decreasing values of our discount factor $\gamma$ reduces the error in our value estimates. As predicted by equation 9 this is due to the effect $\gamma$ has on the variance of our value estimator. The relative residual error is calculated by

$$\text{rel residual error} = \frac{|v_{est} - v_{true}|}{|v_{true}|}, \quad (15)$$

where we normalize over all $N$ states of our environment. We also see that as the sample size increases, relative residual error decreases.

Figure 5 compares the residual error of MCMI to ML and TD. As we predicted MCMI has an accuracy rate comparable to ML and is considerably better than TD. In fact, MCMI is at times more accurate than ML which may be due to the bias ML has at low sample size.

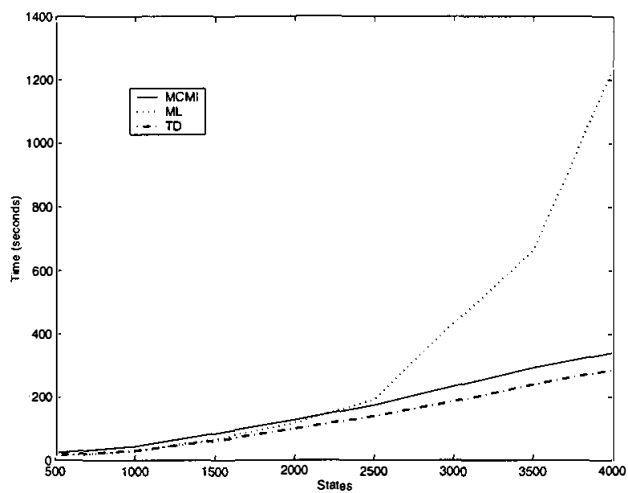

Figure 6: Runtime vs. Number of States (N) for differing solvers, T = 20000

Figure 6 verifies our prediction that MCMI runs at $O(TN)$ just as TD does. Here we fixed the sampling size at 20,000. As we can see, both MCMI and TD run approximately linear to the state space size $N$. ML's worst case $O(N^3)$ time can be seen as the state size becomes very large. At small state size ML can often run faster than TD or MCMI.

Figure 7 uses Boyan's (1999) implementation of LSTD which runs at $O(Tk^2)$ time. We fix both the number of sampling steps at T = 20,000 and the number of feature vector states at $k = 100$. We bound the number of states visited during sampling to a total of $m = 100$. The only varying parameter is the state size. LS-MCMI runs at $O(Tm)$ time. Neither LS-MCMI nor LSTD include $N$ in their runtimes. Therefore, as one would expect, the graph of figure 7 shows the runtimes of both LS-MCMI and LSTD running at approximately constant time.

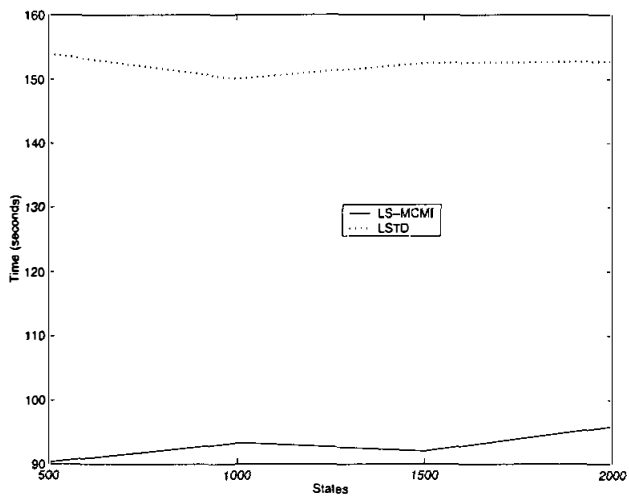

Figure 7: Runtime vs. Number of States (N) for LS-MCMI and LSTD, T = 20000

## 6 CONCLUSION

In this paper we have presented a Monte Carlo matrix inversion value estimation algorithm for policy evaluation. This method possesses the advantages of the speed of the temporal differencing approach, running at $O(TN)$ time when estimating all $N$ states of a system, combined with the superior accuracy of the maximum likelihood approach. Unlike both temporal differencing and the maximum likelihood approach, Monte Carlo matrix inversion value estimation can estimate independently individual states without updating the value estimates of any other states. Therefore the actual runtime for estimating a single state for MCMI is $O(T)$. This model-based method requires model storage of $O(N)$. The state update independence of MCMI allows us to implement a least-squares Monte Carlo matrix inversion (LS-MCMI) algorithm by compressing the system after computing value estimates of states visited during sampling. The LS-MCMI algorithm thus runs in $O(Tm)$ where $m$ is the number of unique states visited.

From our analysis, when performing value estimation for reinforcement learning problems, if storage space is not too much a concern, then the model-based approaches should be preferred. Maximum likelihood value estimation can be used efficiently for low state size and for large state size Monte Carlo matrix inversion may be efficiently applied. For exponentially large state spaces, a least-squares Monte Carlo matrix inversion can be used.

In regards to future work, we will look into policy improvement techniques for MCMI using such methods as policy iteration and value iteration. We may also apply a policy iteration method to our LS-MCMI method and compare it to Lagoudakis and Parr's (2001) least-squares policy iteration technique.